\pdfoutput=1
\documentclass[11pt]{article}
\usepackage[final]{acl}

\usepackage{times}
\usepackage{latexsym}
\usepackage[T1]{fontenc}
\usepackage[utf8]{inputenc}
\usepackage{microtype}
\usepackage{inconsolata}
\usepackage{graphicx}

\title{Assess and Prompt: A Generative RL Framework for Improving Engagement in Online Mental Health Communities}

\author{Bhagesh Gaur$^1$, Karan Gupta$^1$, Aseem Srivastava$^1$, Manish Gupta$^2$, Md Shad Akhtar$^1$ \\
  $^1$IIIT Delhi, India; $^2$Microsoft, India\\
  \texttt{\{bhagesh20558,karan21258,aseems\}@iiitd.ac.in,}\\\texttt{ gmanish@microsoft.com, shad.akhtar@iiitd.ac.in} \\
}

\newcommand{\data}{\textsc{ReddMe}}
\newcommand{\model}{\textsc{MH-Copilot}}
\newcommand{\tx}{\textsc{CueTaxo}}
\newcommand{\mar}[1]{\textcolor{black}{#1}}

\usepackage{todonotes}

\usepackage{graphicx}
\usepackage{amsmath}
\usepackage{soul}
\usepackage{xcolor,colortbl}
\usepackage{bibentry}
\usepackage{booktabs}
\usepackage{multirow}
\usepackage{arydshln}
\usepackage{caption}

\newcolumntype{H}{>{\setbox0=\hbox\bgroup}c<{\egroup}@{}}

\usepackage{enumitem}
\begin{document}

\maketitle
\begin{abstract}
  Online Mental Health Communities (OMHCs) provide crucial peer and expert support, yet many posts remain unanswered due to missing support attributes that signal the need for help. We present a novel framework that identifies these gaps and prompts users to enrich their posts, thereby improving engagement. To support this, we introduce \data, a new dataset of 4,760 posts from mental health subreddits annotated for the span and intensity of three key support attributes: event (\textit{what happened?}), effect (\textit{what did the user experience?}), and requirement (\textit{what support they need?}). Next, we devise a hierarchical taxonomy, \tx{}, of support attributes for controlled question generation. Further, we propose \model, a reinforcement learning-based system that integrates (a) contextual attribute-span identification, (b) support attribute intensity classification, (c) controlled question generation via a hierarchical taxonomy, and (d) a verifier for reward modeling. Our model dynamically assesses posts for the presence/absence of support attributes, and generates targeted prompts to elicit missing information. Empirical results across four notable language models demonstrate significant improvements in attribute elicitation and user engagement. A human evaluation further validates the model's effectiveness in real-world OMHC settings.
\end{abstract}

\begin{figure}
    \centering
    \includegraphics[width=\columnwidth]{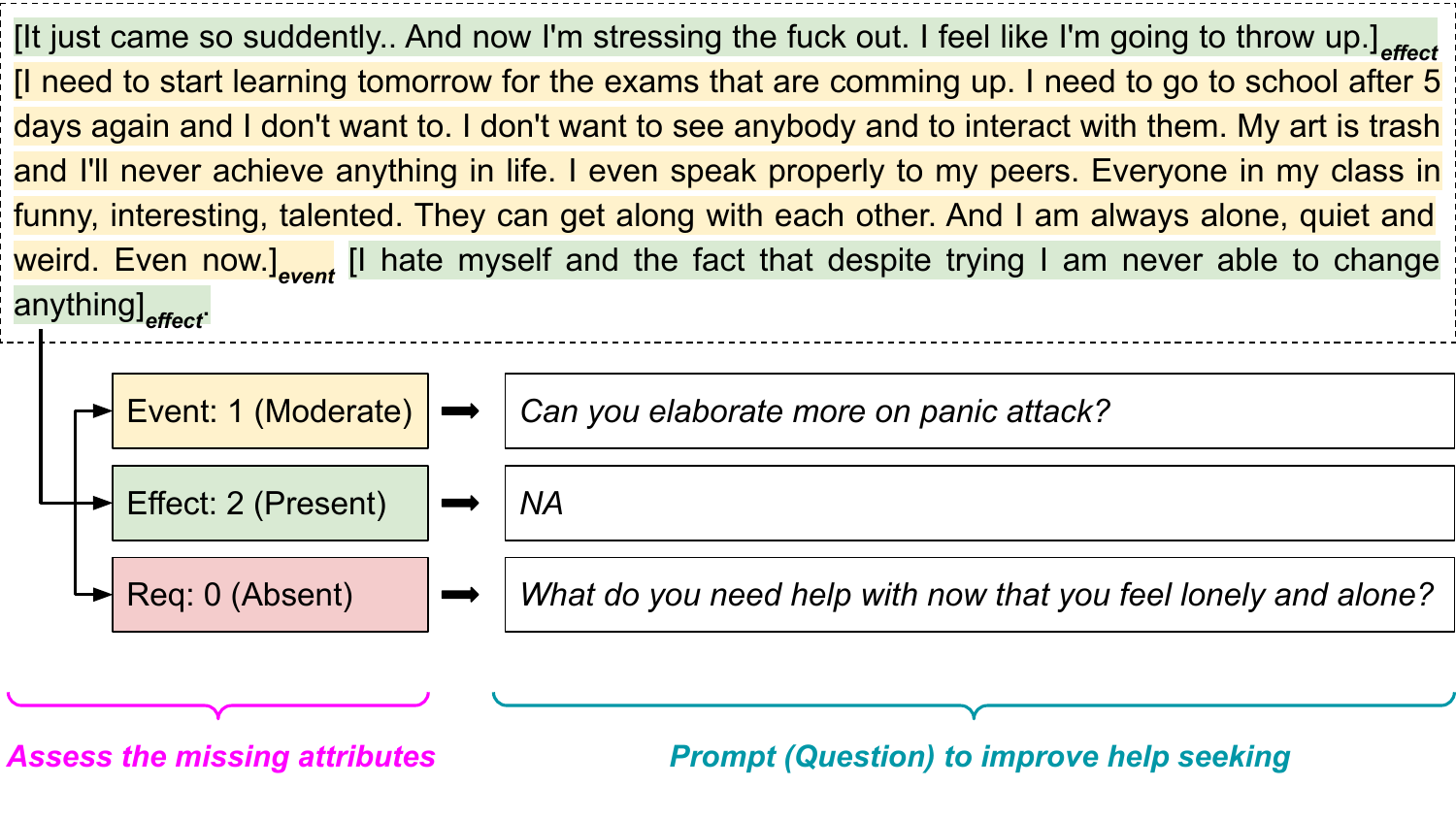}
    \caption{A sample post on a mental health subreddit. Our proposed framework attempts to identify support attributes and then enforce the language model to generate guiding questions.}
    \label{fig:pg1Exam}
    \vspace{-4mm}
\end{figure}

\section{Introduction}
The proliferation of Online Mental Health Communities (OMHCs) has evolved the landscape of the mental health support ecosystem, providing platforms where individuals can share experiences, seek advice, and obtain suggestions or support. These digital arenas have democratized access to mental health resources, especially for those who might otherwise face barriers to traditional support systems, possibly due to stigma, cost, or scarcity of experts \cite{Naslund_Aschbrenner_Marsch_Bartels_2016}. Despite their potential, a significant challenge persists: many support-seeking posts remain unanswered, leaving individuals without the assistance they seek due to poor readability. Several times, user forget to mention essential details of what happened, what did the user experience and what support they need. How do we help users in better articulation of posts to increase the chance of receiving a response?

Previous studies have primarily focused on either assessing community engagement \cite{10.1145/3290605.3300294, Sharma_Choudhury_Althoff_Sharma_2020, info:doi/10.2196/49074} or automating responses \cite{sharma2022humanaicollaborationenablesempathic}. 
\mar{At the same time, a rich body of prior work highlights Reddit’s prominence as a valuable resource for mental health research \cite{Chen_2023,10394671,10.1145/3269206.3271732,rai2024,lokala2022}. 
While language models (LMs) have enhanced support providers’ experiences, as demonstrated by \citet{10.1145/3442381.3450097} in empathy induction, a critical gap remains in directly assisting support seekers. Drawing inspiration from traditional therapeutic stages, building bond, identifying root causes, evaluating consequences, and clarifying goals \cite{adhikary2024exploringefficacylargelanguage,Benjet}, our work translates these paradigms into online settings using state-of-the-art LMs. With better contextual understanding, interpersonal insights \cite{10.1093/jamia/ocad071}, and explainable AI \cite{ibrahimov2024}, we aim to bridge the gap by focusing on improving support seekers’ experiences in OMHCs.} 

To address these gaps systematically, we first curate a novel first-of-its-kind dataset, \data, extending the publicly available mental health corpus -- BeCOPE \cite{criticalBehaviorAseem}. \data\ contains $4760$ posts, where each post is annotated for the spans and presence/absence of three pivotal support seeker's cues: {\em event, effect,} and {\em requirement}. Next, we also propose a dedicated taxonomy of support attributes, \tx, that allows controlled prompting for LMs and is responsible for question generation. Finally, we propose \model, a reinforcement learning (RL)-based framework tailored to enhance the posting behavior of support seekers on OMHCs. 
\model\ operates on two principles: (a) assessing the support-seeker's post for missing support attributes 
and (b) prompting a question to better articulate their posts by using \tx. Figure \ref{fig:pg1Exam} illustrates an example of how a user can be prompted for different support seekers' cue intensity levels. A reward model ensures the quality and relevance of these generations through RL.

\begin{table*}[t]
  \begin{minipage}[b]{0.34\textwidth}
    \centering
    \includegraphics[width=\textwidth]{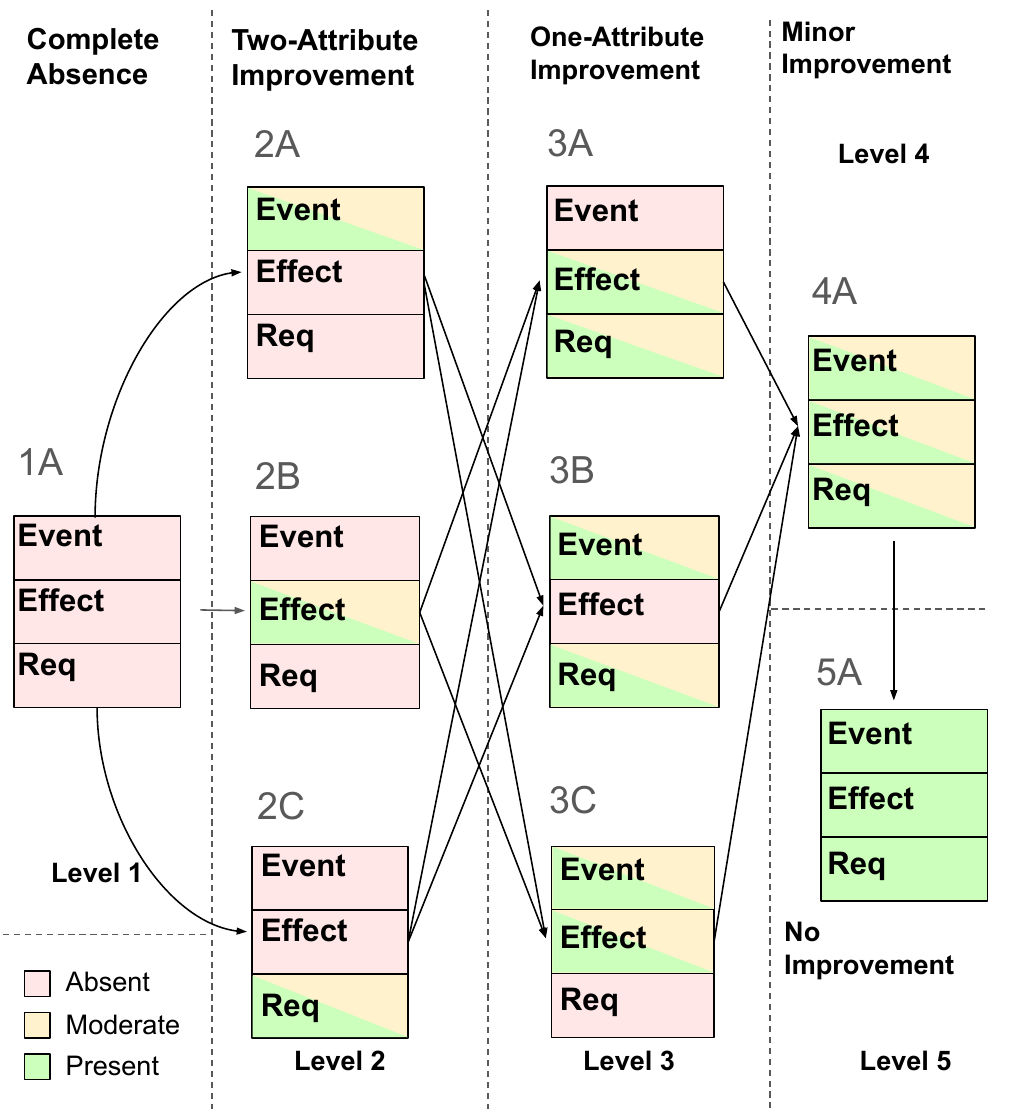}
    \captionof{figure}{Taxonomy: \tx{}.}
    \label{fig:taxonomy}
  \end{minipage}
  \hfill
  \begin{minipage}[b]{0.65\textwidth}
    \centering
    \tabcolsep1pt
    \scriptsize
    \begin{tabular}{l|p{0.31\textwidth}|p{0.31\textwidth}|p{0.31\textwidth}}
\hline
~ & \bf Event & \bf Effect & \bf Requirement \\ \hline
        \bf 1A & Can you tell me what happened?You can be as specific as you like. & Could you describe the specific effect the event has had on you?& What kind of support or help you feel would be most beneficial?\\ \hline
        \bf 2A & Can you elaborate more on X?& How did X make you feel?& What do you need help with now that X?\\ \hdashline 
        \bf 2B & What made you feel X?&Can you elaborate more on X?& What can help you overcome X?\\ \hdashline 
        \bf 2C & What happened that you want X?& Why are you wanting X?\newline What caused you to need X?& Can you elaborate more on X?\\\hline
        \bf 3A & What made you feel X?\newline What happened that you want X?& Can you elaborate more on X?& Can you elaborate more on X?\\  \hdashline
        \bf 3B & Can you elaborate more on X?& How did X make you feel?\newline Why are you wanting X?\newline  What caused you to need X?& Can you elaborate more on X?\\ \hdashline
        \bf 3C & Can you elaborate more on X?& Can you elaborate more on X?& What do you need help with now that X?\newline What can help you overcome X?
        \\\hline
        \bf 4A & Can you elaborate more on X?& Can you elaborate more on X?& Can you elaborate more on X?\\ \hline
\end{tabular}
\captionof{table}{{Taxonomy-based question prompts} annotated for each taxonomy level. X signifies a user-mentioned entity. Questions are generated only for moderate or absent intensities; hence level 5A is not shown.}
\label{tab:questions}
    \end{minipage}
\end{table*}

We benchmark \model\ against four LLMs: Llama-3~\citep{llama3}, Mistral~\citep{mistral}, Phi-3~\citep{phi3}, and Gemma-2~\citep{gemma2}. 
Our proposed model demonstrates consistent improvements across all baselines, with a notable improvement of $+27.49\%, +17.80\%$, $+2.81\%$ and $+17.54\%$ on ROUGE-L, BLEU-4, BERTScore and METEOR respectively. We further conduct human evaluation and qualitative analysis to assess the quality of generations. Our contributions are summarized below:

\begin{itemize}[noitemsep,leftmargin=*,nolistsep=1]

    \item We propose a \textbf{novel problem} of enhancing support-seekers’ posting behavior using linguistic and contextual cues to identify missing attributes, prompting users to share more.

    \item A \textbf{novel dataset} containing mental health-related posts from Reddit, marked with key support attributes at span and intensity levels.

    \item A \textbf{reinforcement-learning-based framework} to assess support attributes and generate taxonomy-guided questions to enhance post responsiveness.

    \item Evaluation against notable LMs, where our approach consistently outperformed on the complete suite of evaluation metrics. 
\end{itemize}

\paragraph{Reproducibility.} The \data\ dataset\footnote{\url{https://huggingface.co/datasets/Shadowking912/REDDME}} and code for \model\footnote{\url{https://github.com/flamenlp/MH-COPILOT}} is open sourced for research purposes.

\section{Related Work}
We study the support-seeking behavior on OMHCs. \mar{The study of psychological aspects of human behaviour and counselling dates long back \cite{eliza96}}. We aim to study these aspects for controlled question generation. 

\paragraph{Online Mental Health Communities (OMHCs) and Support Provision.}
The increasing reliance on digital platforms for mental health support highlights the need to enhance user experience in OMHCs. \citet{info:doi/10.2196/49074} and \citet{althoff-etal-2016-large} developed tools to improve interactions between support seekers and peer supporters. Despite progress, direct assistance for support seekers remains limited, a gap our work addresses. Studies have documented barriers to offline care \cite{olfsonCapacity} and stigma in help-seeking \cite{10.1093/her/16.6.693}, underscoring the critical role of peer support. While \citet{10.1145/3442381.3450097} and \citet{10.1145/3290605.3300261} have advanced empathic communication, others like \citet{info:doi/10.2196/12529} and \citet{zhang-etal-2019-finding} have worked on moderation and engagement. In contrast, our research focuses on enhancing responsiveness to support-seeking posts through controlled prompting.

\paragraph{Controlled Prompting in LLMs.} Controlled prompting is essential for refining outputs from LLMs in precision-sensitive contexts \cite{evuru-etal-2024-coda}. While LLMs handle vast data, studies show LLMs often struggle with systematic generalization, producing content that is both coherent and high-quality \cite{petruzzellis2024benchmarkinggpt4algorithmicproblems} . 

\paragraph{Reward Modeling and Self-Improvement in Language Modeling.} In sensitive domains like OMHCs, accuracy and appropriateness are vital. To gain more control, we incorporate self-improvement techniques \cite{10.5555/3600270.3601396} that refine model outputs during fine-tuning. We also use Direct Preference Optimization (DPO) ~\cite{DPO} to guide the model toward generating more context-sensitive questions. This is particularly important when handling deviations from standard inputs \cite{hosseini2024vstartrainingverifiersselftaught}. 
By integrating controlled prompting and reward modeling, our framework enhances output quality while aligning better with the needs of mental health support seekers, addressing a key research gap.

\section{Dataset: \data}
\label{dataset}
We propose \data, a manually annotated corpus extended from the publicly available mental health subreddit corpus, BeCOPE  \cite{criticalBehaviorAseem}. 
BeCOPE categorizes posts into three primary categories: (a) \textit{interactive}, if there are back-and-forth conversations between the OP (original poster) and peers (b) \textit{non-interactive}, if the post engages peers, but the OP does not reply to peers (c) \textit{isolated}, if the post does not have any comments. 
We select $4,760$ posts and manually annotate them with support attributes: {\em event, effect,} and {\em requirement} on span (rationale) and intensity level.  

\begin{figure}[b]
\includegraphics[width=0.45\textwidth]{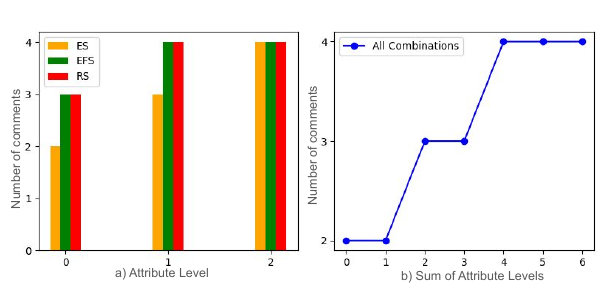}
    \caption{(a) Distribution of median number of comments for each level of the support attribute, (b) Median Number of comments for sum of level of event effect and requirement for a post.}
    \label{fig:attr_analysis}
\end{figure}

\paragraph{Motivation for Attribute Selection.} During a conversation with support seekers, therapists often provide support in stages over time \cite{betterhelp_steps_2025}. First, they build rapport with the patient, which is essential for creating a safe and trusting environment \cite{adhikary2024exploringefficacylargelanguage}. Following this, they try to identify the root cause of the patient's condition, which could be a single \textbf{event} or a series of events, potentially involving other individuals \cite{social-determinants,personality}. Next, the therapist evaluates the potential \textbf{effects} of these events on the patient, which can range from depression and suicidal thoughts to changes in appetite or sleep \cite{Benjet}. It's important to note that this evaluation is subjective, and the effects may not always be apparent \cite{Tiwari2020}. Finally, the therapist identifies what the patient seeks (\textbf{requirement}) from the therapy, which helps keep them proactively involved and committed \cite{10.1093/intqhc/mzae009}. 
It's crucial for understanding the support seeker's perspective \cite{doi.org/10.1002/cpp.2479,saxena2022} before initiating therapy. Further, Figure \ref{fig:attr_analysis} shows an increase in the number of comments received by the support seeker as the level of each attribute increases.

\paragraph{Support Attributes.} Building up on these theories from previous work, we define our three primary support attributes, which are critical to capture the help-seeking nature of posts: \textit{event}, \textit{effect}, and \textit{requirement}. Additional details of attributes are in Appendix \ref{appx:attributes}.

\begin{itemize}[noitemsep,leftmargin=*,nolistsep=1]
\item {\bf Event} encapsulates the specific situation, activity, or event that is the focal point of the support seeker’s concern. The explicit detailing of such events provides a contextual background essential for overall background understanding, as suggested by \citet{sharma2020computational}. 

\item {\bf Effect} targets the impact or consequences of the identified event from the support seeker. By elucidating the effect, the post conveys the emotional or practical repercussions of the situation, thereby inviting more targeted and empathetic responses. 

\item {\bf Requirement} lays out the expectation (e.g., informational support, instrumental aid, etc.) of support seeker from peers. It is crucial in directing the nature of the assistance sought, and thereby guiding the potential support trajectory. This aligns with the insights laid out by \citet{sharma2022humanaicollaborationenablesempathic}, highlighting the importance of clearly articulated needs for effective support. 
\end{itemize}

\begin{table*}[t]\centering
\scriptsize
\begin{tabular}{lccccccccccccccc}
\toprule
\multirow{3}{*}{\bf Split} & \multirow{3}{*}{\bf \#Posts} & \multirow{3}{*}{\bf \#Prompts} & \multirow{3}{*}{\bf APoL} & \multicolumn{12}{c}{\bf Support Attributes}\\ 
\cmidrule{5-16}
~ & ~ & ~ & ~ & \multicolumn{3}{c}{\bf Event} & \bf AEvL & \multicolumn{3}{c}{\bf Effect}& \bf AEfL & \multicolumn{3}{c}{\bf Requirement} &\bf  AReL \\
\cmidrule{5-7} 
\cmidrule{9-11} 
\cmidrule{13-15}
~ & ~ & ~ & (words) & Ab & Mo & Pr & (words) & Ab & Mo & Pr & (words) & Ab & Mo & Pr & (words) \\ 

\midrule

{\bf Train} & 3331 & 5533 & 180.99 & 938 & 568 & 1825 & 66.35 & 1531 & 415 & 1385 & 26.50 & 1722 & 359 & 1250 & 19.62 \\ 

{\bf Val} & 953 & 1583 & 168.25 & 268 & 162 & 523 & 61.31 & 438 & 118 & 397 & 26.72 & 493 & 104 & 356 & 18.51 \\ 

{\bf Test} & 476 & 793 & 192.81 & 134 & 82 & 260 & 69.51 & 221 & 58 & 197 & 25.89 & 246 & 52 & 178 & 18.82 \\
\midrule
{\bf Total} & 4760 & 7909 & 179.62 & 1340 & 812 & 2608 & 65.70 & 2190 & 591 & 1979 & 26.48 & 2461 & 515 & 1784 & 19.31\\


\bottomrule
\end{tabular}
\caption{Statistical Analysis of \data. Train, val, test split is 70:20:10. \textbf{APoL}, \textbf{AEvL}, \textbf{AEfL}, and \textbf{AReL} are the average lengths of \textit{post}, \textit{event span}, \textit{effect span}, and \textit{requirement span}, respectively. \textbf{\#Prompts} defines number of question-prompts. \textbf{Ab:} Absent, \textbf{Mo:} Moderately Present, \textbf{Pr:} Present.}
\label{tab:datasetStats}
\end{table*}

\paragraph{Annotation Process.}
The overall annotation process is crafted on three major fronts: (a) rationale (span-level) annotation considering the three support attributes, (b) intensity level annotation, and (c) taxonomy-based prompt question annotation. 

\paragraph{Rationale Annotation.} For each support attribute, we annotate spans corresponding to the event, effect, and requirement attributes by bounding them with special $\langle \text{\em start} \rangle$ and $\langle \text{\em end} \rangle$ tokens -- we use $\langle \text{\em es} \rangle$, $\langle \text{\em ee} \rangle$ for event, $\langle \text{\em efs} \rangle$, $\langle \text{\em efe} \rangle$ for effect, and $\langle \text{\em rs} \rangle$, $\langle \text{\em re} \rangle$ for requirement, respectively. 
Due to the informal nature of the Reddit posts (incorrect grammar, inconsistent punctuation, absence of sentence markers, etc.), we execute the rationale annotation on a token level, with a preference for marking the complete sentences as much as possible. We share additional details pertaining to the rationale in Appendix \ref{appx:additional rationale}.  

\paragraph{Intensity Annotation.} To quantify the presence of support attributes, we employ a three-scale Likert rating system for annotation: {\em absent (0)}, {\em moderately-present or `moderate' (1)}, and {\em well-described or `present' (2)}, indicating the degree to which each attribute is reflected in the original post. For instance, if a post does not carry any {\em event} rationale, the corresponding intensity label is marked as {\em absent}. An example is shown in Figure \ref{fig:pg1Exam}.

\paragraph{Taxonomy-based Reference Question Annotation.} We introduce a custom taxonomy, \tx, that enables a structured and interpretable approach to prompting users with targeted questions for improving support attributes. As shown in Figure \ref{fig:taxonomy}, \tx\ consists of five levels, each representing varying degrees of completeness for the three support attributes within a post. Level $1$ (none present; red-colored), Level $5$ (all well-described; green-colored), and Levels $2$-$4$ capturing partial expression (yellow-colored).
To make our taxonomy generalizable, we have similar questions for the same intensities. For cases with absent attribute information, our evaluations reveal the need for customized questions that utilize context from other attributes in the follow-up suggestions for the support seekers.
As a result, \tx\ assists in the selection of the ideal template question, as depicted in Table \ref{tab:questions}.
For instance, in the case where {\em event} is missing, we ask annotators to complete: {\em What happened that you wanted X?}. $X$ is a placeholder that annotators can fill in with the help of rationales. 

\paragraph{Annotator Details.}
We employ two annotators, aged 21–24, with expertise in linguistics and the relevant domain, to annotate the dataset. 
Additionally, an expert reviewer acts as a moderator and periodically evaluates the annotation quality. Both annotators are provided with guidelines and examples to support the consistent identification and categorization of each annotation type within the posts, which we present in the Appendix \ref{appx:annotator guidelines}. During the initial training phase, annotators participate in regular calibration sessions and discussions to align their interpretation of the guidelines and resolve any annotation discrepancies. Following the training phase, the annotators independently annotated two equally divided subsets of the dataset. To assess inter-annotator agreement, we calculated Cohen’s Kappa scores on a random sample of 50 instances. The resulting agreement scores were $0.885$ for event, $0.886$ for effect, and $1.000$ for requirement, indicating a high degree of consistency. 

Detailed statistics of \data\ are shown in Table \ref{tab:datasetStats}. Further dataset details are in Appendix \ref{appx:subreddits}.

\begin{figure*}[t]
    \includegraphics[width=1\textwidth]{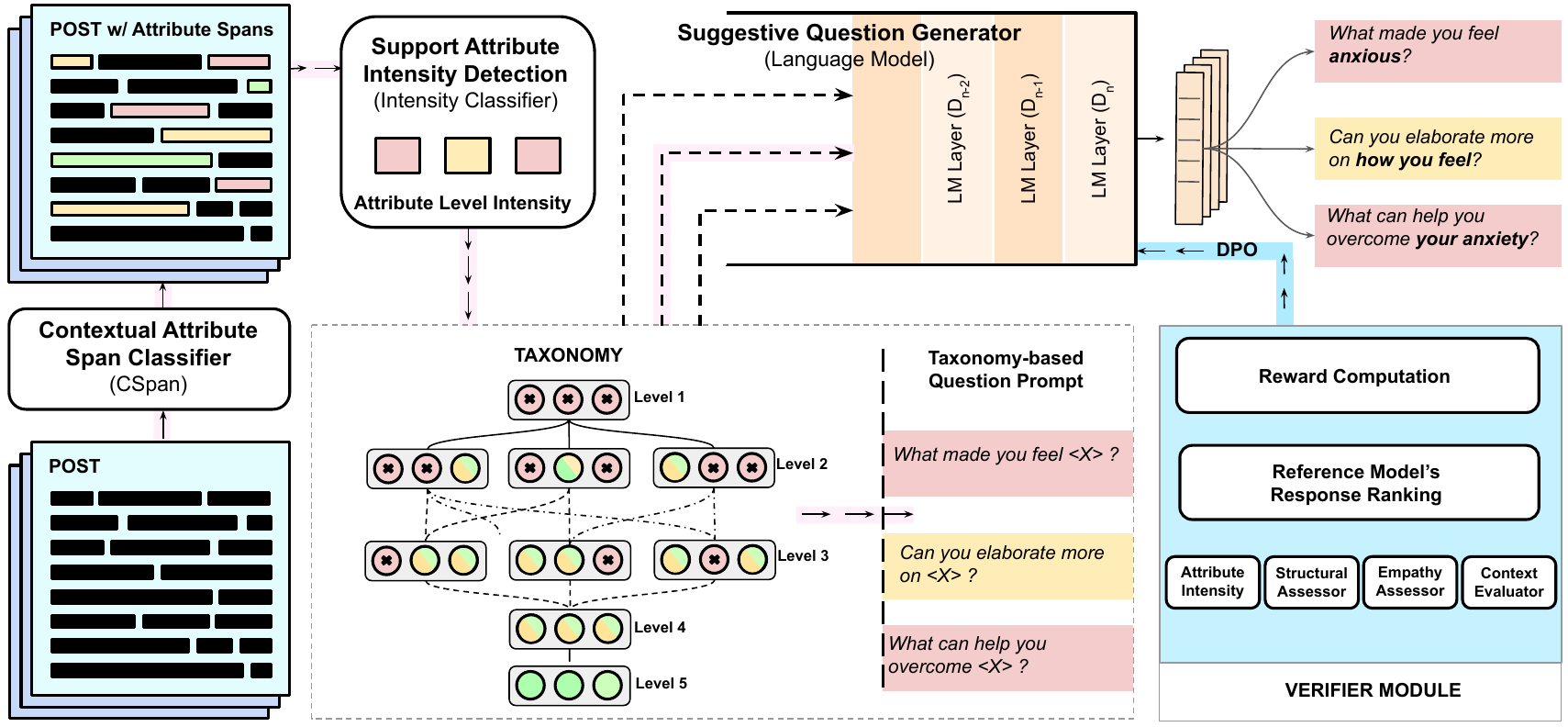}
    \caption{\model\ Framework: The four contributory modules are (a) contextual attribute span classifier, (b) support attribute intensity detection, (c) taxonomy-based question generation, and (d) verifier. }
    \label{fig:model}
\end{figure*}

\section{Methodology}
In this work, we propose \model, a novel framework that prompts support seekers with guiding questions based on their posts to elicit improvements. Our framework first identifies the presence of such attributes and further generates the required prompts via a hierarchical taxonomy, \tx. To achieve this, we divide \model\ into four fundamental modules: (a) a contextual attribute-span identifier (CSpan), (b) a support attribute intensity classifier, (c) a guiding question generator, and (d) a verifier. A schematic diagram of framework is shown in Figure \ref{fig:model}.  Next, we discuss each of these modules in detail.

\paragraph{Contextual Span Identifier (CSpan).}
The CSpan module is responsible for identifying and extracting the relevant contextual spans that represent key attributes of a support-seeking post: {\em event, effect,} and {\em requirement}. These components provide essential context that can help responders better understand the situation described by the support seeker. We frame this as an entity extraction task and fine-tune RoBERTa \cite{liu2019robertarobustlyoptimizedbert}
on the \data\ dataset to extract support attribute spans. 
Precisely, for each post $P$ in the dataset, the input consists of tokens $\{t_1,t_2, \dots,t_n\}$
and corresponding annotated labels $\{a_1,a_2,\dots,a_n\}$,
where each label could be one of event, effect or requirement. 
We utilize these spans further in the identification of the relevant attribute's intensity.

\paragraph{Support Attribute Intensity Classifier.}
After contextual span identification, the next step is to assess the intensity of each attribute, as a typical multi-class, multi-label classification problem. The classifier takes the token-level span outputs from the $CSpan$ module as input and yields an intensity vector, $V = \{ v_\text{event}, v_\text{effect}, v_\text{req} \}$, where each $v_i \in \{ 0, 1, 2\}$ signifies absent ($0$), moderate ($1$), or present ($2$), respectively. Subsequently, these values act as the degree to which the support attributes are articulated in the post. We fine-tune RoBERTa for this task, which outperforms three baselines (Table \ref{tab:intensityresults}). The output of this module guides the question-generator module by providing information on how well the attributes are expressed using our taxonomy.

\paragraph{Guiding Question Generator.} We embed the \tx\ taxonomy in the language model via prompting (Appendix \ref{appx:prompts}), by adding the levels of all the attributes in the prompt for the language model, enabling it to generate questions that are informed by the identified spans and their corresponding intensity levels. It ensures that generated prompts abide by the support attributes. We perform supervised fine-tuning of LLMs on \data\ to train {\em question generator} to learn the patterns of generating helpful prompts that align with the taxonomy. Given our prompt, we expect the output to be contextually relevant questions for each attribute. For example, if the event component is moderately described, the prompt generator generates a question such as, {\em ``Can you describe more about the event that led to your current feelings?''}. Next, we utilize a reward model to control the generation via DPO \cite{DPO}.

\paragraph{Verifier and Reward Modeling.} 
Verifier module validates the outputs of the question generator module across these dimensions: 

\begin{itemize}[noitemsep,leftmargin=*,nolistsep=1]
\item {\bf Support Attribute Category Classifier.} This multi-class classifier verifies whether each generated question aligns with the intended support attribute. We employ the best-performing baseline, RoBERTa, as a reference model in the verifier model. If the question correctly matches the target attribute, it receives a reward score of 1; otherwise, it receives a reward score of 0.
\item {\bf Contextual Grounding Evaluator.} The context map component uses a reference language model to evaluate how closely the generated question aligns with the original post. Considering the large and complex contexts, we employ Llama-3-70b \cite{llama3} for this purpose. A score between 0 and 1 is assigned, with higher scores indicating stronger contextual relevance.
\item {\bf Structural Adherence Assessor.} We ensure that the generations adhere to a pre-defined structure, defined within the taxonomy template. A score of $1$ is for correct format and $0$ for deviations.
\item {\bf Empathy Assessor.} We ensure that the generations are empathetic toward the support seeker, and carefully avoid triggering language or tone. We use Llama-3-70b to generate a score between 0 and 1; higher scores indicate higher empathy. 
\end{itemize}

\paragraph{Reward Modeling.}The final reward ($r$) for each sample is calculated by combining the scores from all reward parameters, as shown in Equation \ref{eq:reward}.
\begin{equation}
    r = \sum_{i \in [1,2,3]} r_{Q_i}^{CC} \cdot r_{Q_i}^{CG} \cdot r_{Q_i}^{EA} \cdot r_{SA}
    \label{eq:reward}
\end{equation}
where \( {Q_1}, {Q_2}, {Q_3} \) represent the questions for event, effect, and requirement components, and \( r^{CC}, r^{CG}, r^{EA}, r^{SA} \) are the scores for category correctness using attribute classifier, contextual grounding using context map, empathy accessor and structure adherence using structure assessor, respectively. We augment preference data to train \model\ with DPO by generating two responses from variant with taxonomy enabled with supervised fine-tuning (SFT+\tx) for the training dataset. These responses are then passed to the verifier module to get a reward score for each of them, $r_1$ and $r_2$ using Eq. \ref{eq:reward}. 

Response with the higher and lower rewards are chosen as preferred ($y_p$) and non-preferred ($y_{np}$) responses for DPO respectively.
This ranking contains the nuances of the context scores, structure scores, and empathy scores. We create the preferential dataset for the responses as: $ \mathcal{D}=\{x^{(i)},y_p^{(i)},y_{np}^{(i)}\}$.
Here, $x^{(i)}$ is the prompt for generating $\langle y_p^{(i)},y_{np}^{(i)} \rangle$. The resulting preferential dataset is utilized for reward tuning using the likelihood function shown in Eq.~\ref{eq:dpolikelihood}. Hinge-loss is used as shown in Eq.~\ref{eq:hinge}, where $\sigma$ denotes the logistic function, and $\beta$ is a scaling parameter. 
\begin{equation}
    F=\beta \log \frac{\pi_\theta(y_p|x)}{\pi_{ref}(y_p|x)}- \beta \frac{\pi_\theta(y_{np}|x)}{\pi_{ref}(y_{np}|x)}
    \label{eq:dpolikelihood}
\end{equation}
\begin{equation}
\mathcal{L}_{DPO}(\pi_\theta;\pi_{ref})=-E_{(y_p,y_{np}) \sim D} [\log \sigma (F)]
\label{eq:hinge}
\end{equation}

\begin{table}[!b]
\centering
\scriptsize
\tabcolsep4pt
\begin{tabular}{llcccc}

\toprule
Models &{Accuracy} & Precision & Recall & F1 \\
\midrule
        
T5-Large \cite{2020t5} & 66.89 & 12.54 & 36.36 & 18.65 \\ 
SpanBERT \cite{joshi-etal-2020-spanbert} & 69.59 & 22.70 & 33.66 & 27.12 \\ 
RoBERTa-base & \textbf{74.95} & 37.33 & 60.47 & 46.16 \\
RoBERTa-large & 73.00 & \textbf{45.29} & \textbf{62.50} & \textbf{52.52} \\
        
\bottomrule
\end{tabular}
\caption{Span prediction results from CSpan.}
\label{tab:spanresults}
\end{table}

\begin{table}[!b]
\centering
\scriptsize
\tabcolsep4pt
\begin{tabular}{lccccc}
\toprule
\multirow{2}{*}{Models} & \multicolumn{2}{c}{Accuracy} 
 & & \multicolumn{2}{c}{F1}\\ \cmidrule{2-3} \cmidrule{5-6}
& ORD & MSE & & ORD & MSE \\ \midrule
        BERT \cite{BERT} & 84.00 & 82.30 && 72.00 & {\bf 73.10} \\
        MentalBERT \cite{mentalBERT}& 85.70 & {\bf 83.70} && 73.20 & 68.10 \\
        MentalRoBERTa & 84.80 & 83.20 && 71.20 & 68.30\\
        RoBERTa & \textbf{86.60} & 83.30 && \textbf{77.40} & 71.80  \\ 
\bottomrule
\end{tabular}
\caption{Attribute intensity classifier results.}
\label{tab:intensityresults}
\end{table}

\begin{table*}[ht]\centering
\centering
\scriptsize
\tabcolsep2pt
\begin{tabular}{lp{1.5cm}cccccccccccc}
\midrule
\bf \multirow{2}{*}{Models} & & \multirow{2}{*}{R1} 
& \multirow{2}{*}{R2} & \multirow{2}{*}{RL} & \multirow{2}{*}{B1} & \multirow{2}{*}{B2} & \multirow{2}{*}{B3} & \multirow{2}{*}{B4} & \multicolumn{3}{c}{BERTScore}&\multirow{2}{*}{METEOR}\\
\cmidrule{10-12}
& & & & & & & & &P & R & F1& \\\midrule

Mistral \cite{mistral} & \multirow{4}{*}{\bf Zero Shot} & 49.15 & 37.47 & 47.58 & 50.69 & 45.52 & 40.95 & 35.68 & 87.01 & 89.77 & 88.36 & 61.63\\ 
Phi-3 \cite{phi3}  & & 39.22 & 26.98 & 36.29 & 43.80 & 40.04 & 37.48 & 34.46 & 85.28 & 89.47 & 87.32 & 50.09 \\ 
Llama-3 \cite{llama3} & & 43.60 & 28.00 & 40.80 & 50.10 & 45.30 & 42.20 & 38.61 & 87.00 & 89.12 & 88.00 & 48.50\\ 
Gemma-2  \cite{gemma2} & & 45.24 & 31.64 & 42.03 & 52.22 & 48.23 & 45.48 & 42.41 & 87.27 & 89.96 & 88.58 & 61.51\\ 
\midrule

Mistral \cite{mistral} & \multirow{4}{*}{\bf SFT} & \underline{72.39} & \underline{62.25} & \underline{69.71} & 82.25 & \underline{80.00} & \underline{78.12} & \underline{76.21} & \underline{96.55} & \underline{95.58} & \underline{96.04} & \underline{79.83}\\ 
Phi-3 \cite{phi3} & & 66.88 & 56.66 & 64.14 & 78.97 & 76.48 & 74.57 & 72.69 & 96.07 & 94.53 & 95.25 & 75.65 \\ 
Llama-3 \cite{llama3} & & 71.30 & 61.30 & 68.40 & \underline{82.30} & 79.70 & 77.82 & 75.91 & 96.30 & 95.30 & 95.80 & 79.00 \\ 
Gemma-2  \cite{gemma2} & & 68.20 & 58.04 & 65.58 & 80.39 & 77.87 & 75.95 & 74.00 & 96.10 & 94.94 & 95.48 & 76.98\\ 
\midrule
\rowcolor{yellow!34}\bf \model ~ & {\bf SFT + \tx{} + Rew}  & 89.30 & 84.50 & 88.88 & 93.84 & 92.36 & 91.12 & 89.78 & 98.81 & 98.68 & 98.74 & 93.84\\
    
\midrule
\multicolumn{2}{l}{$\Delta_{\model-{\text{SFT}}}(\%)$} &
\textcolor{blue}{$\uparrow 23.35$} & \textcolor{blue}{$\uparrow 35.74$} & \textcolor{blue}{$\uparrow 27.49$} & \textcolor{blue}{$\uparrow 14.02$} & \textcolor{blue}{$\uparrow 15.45$} & \textcolor{blue}{$\uparrow 16.64$} & \textcolor{blue}{$\uparrow 17.80$}& \textcolor{blue}{$\uparrow 2.34$} & \textcolor{blue}{$\uparrow 3.24$} & \textcolor{blue}{$\uparrow 2.81$} & \textcolor{blue}{$\uparrow 17.54$} \\ 
\midrule
\end{tabular}
\caption{Performance comparison of the proposed method, \model, with four LLMs in two settings, including \textbf{zero-shot} and supervised-finetuning (\textbf{SFT}). Our proposed version, \textbf{\model}, 
is built upon Gemma-2, 
encompasses controlled taxonomy-based generations (\textbf{\tx{}}), and is tuned using reward modeling (\textbf{Rew}). The last row presents our approach's improvement (in \%) with the best variant in the SFT category instead of the zero-shot category for a fair comparison. We employ standard generation metrics: \mar{ROUGE (R-1, R-2, and R-L), Meteor, BLEU (B-1, B-2, B-3, and B-4), and BERTScore (BS)}.}
\label{tab:results}
\end{table*}

\section{Experiments, Results, and Analyses}
In this section, we discuss our obtained results from a single run and the effect of each module in \model, followed by human evaluation and analysis. Additional details are shown in Appendix~\ref{appx:experimentDetails}.

\begin{table}[!b]
\centering
\resizebox{\columnwidth}{!}{
\begin{tabular}{l ccccccccc}
\toprule
\bf Models & R1 & R2 & RL & B1 & B2 & B3 & B4 & BS-F1 & MET\\
 \midrule
\textbf{\model} & 89.30 & 84.50 & 88.88 & 93.84 & 92.36 & 91.12 & 89.78 & 98.74 & 93.84\\
{$-$Rew} & 88.32 & 82.97 & 87.67 & 93.00 & 91.25 & 89.83 & 88.32 &  98.52 & 93.33\\
{$-$Rew$-$\tx{}} & 68.20 & 58.04 & 65.58 & 80.39 & 77.87 & 75.95 & 74.00  & 95.48 & 76.98 \\ 
\midrule
Mistral (SFT) & 72.39 & 62.25 & 69.71 & 82.25 & 80.00 & 78.12 & 76.21 & 96.04 & 79.83\\ 
$+$\tx{} & 88.98 & 83.77 & 88.31 & 93.49 & 91.77 & 90.37 & 88.88  & 98.68 & 93.57 \\ 
$+$\tx{}$+$Rew & 88.14&	82.24&	87.42&	92.49&	90.50&	89.00&	87.52&	98.50&	92.60 \\ 
\midrule
Phi-3 (SFT) & 66.88 & 56.66 & 64.14 & 78.97 & 76.48 & 74.57 & 72.69  & 95.25 & 75.65\\  
$+$\tx{} & 86.80 & 80.94 & 85.98 & 92.04 & 90.10 & 88.69 & 87.10  & 98.33 & 92.16\\
$+$\tx{}$+$Rew & 86.56 & 80.60 & 85.66 & 92.00 & 90.10 & 88.53 & 87.00  & 98.25 & 91.88\\  
\midrule
Llama-3 (SFT) & 71.30 & 61.30 & 68.40 & 82.30 & 79.70 & 77.82 & 75.91  & 95.80 & 79.0 \\   
$+$\tx{}& 87.46 & 81.96 & 86.83 & 92.49 & 90.70 & 89.26 & 87.71  & 98.28 & 94.70 \\
$+$\tx{}$+$Rew & 88.28  & 82.93 & 87.70 & 93.17 & 91.41 & 89.98 & 88.48  & 98.35 &93.23\\  
\bottomrule
\end{tabular}}
\caption{{\bf Ablation Study.} We demonstrate the effectiveness of each model component: taxonomy (\textbf{\tx{}}), and verifier for reward modeling (\textbf{Rew}).}
\label{tab:ablation}
\end{table}

\paragraph{Performance Comparison and Ablation.}
\label{sec:performanceComparison}
We present the performance comparison of our proposed framework, \model, followed by an ablation study with all the contributory modules of \model. Table \ref{tab:spanresults} presents the performance of {\em CSpan} module for the span prediction task. Evidently, RoBERTa-large, outperformed other encoder models on three out of four metrics. It achieves a score of $45.29\%$, $62.50\%$, and $52.52\%$ on precision, recall, and F1, respectively. These scores observe a clear increment of $+21.32\%$, $+3.35\%$, and $+13.77\%$ as compared to the second-best method. On the other hand, Table \ref{tab:intensityresults} shows the performance comparison of {\em intensity-classifier}.  Here, considering the ordinality in labels, we experiment with ordinal loss and MSE loss to achieve a best F1 of $77.4\%$.

For the final task, generating questions, we experiment with a series of LMs. First, in order to select the base LM of \model, we observe zero-shot results as presented in Table~\ref{tab:results}. Evidently, Gemma-2 outperformed Llama-3, Mistral, and Phi-3 across the complete suite of metrics with a notable BERTScore (F1) of $88.58$ and RL score of $42.03$. We reassess their performance in a supervised-finetuning (SFT) setup and observe a mixed behavior, where Llama-3 and Mistral yield competing scores. 
However, we utilize Gemma-2 as our foundational model in \model\ due to its superior quantitative and qualitative performance. 
Finally, \model\ achieves improved scores across all evaluation metrics, with improvement in the range of $+2.34$\% to $+35.74\%$. Last row of Table \ref{tab:results} reports percent improvement of \model\ against the best SFT baseline. 

\begin{table*}[t]
\centering
\scriptsize
\resizebox{\textwidth}{!}{%
\begin{tabular}{p{58em}} \toprule
{\bf Original Post:} {\em \textcolor{magenta}{So I have been taking adderall 10mg for a little over a week and I've been testing it. Today i decided to take adderall and coffee at the same time.} \textcolor{blue}{I think this was a bad idea because i felt extremely jittery and anxious, but the worst part was i came down from them at the same time. I am not fully sure of my personal side effects of adderall yet but coming down at the same time from both I had this weird feeling of nausea, weak, bad headache, tired, extremely zoned out, and almost like a not knowing what's going on kinda feeling, also i'm just getting over the flu. Not sure why this happened but it sucked.} \textcolor{black!30!cyan}{Should I just avoid caffeine for now on, I can do that cause the adderall is doing way more than the caffeine was.} \textcolor{magenta}{The reason i'm asking is cause I couldn't find anyone having this problem.} \textcolor{black!30!cyan}{So if you guys got any tips that would be helpful, like if its the flu, or the adderall, or the caffeine, or if anyone else has experienced this.}} \vspace{2mm}\\
$\Rightarrow$ {\bf Annotated:} $\quad \langle$ Event Level: Moderate $\rangle \quad \langle$ Effect Level: Well-described $\rangle \quad \langle$ Requirement Level: Well-described $\rangle$\\
$\langle$ {\bf Event:} Can you elaborate more on why you are taking adderall? $\rangle$ ; $\langle$ {\bf Effect:} n/a $\rangle$ ; $\langle$ {\bf Requirement:} n/a $\rangle$ \vspace{2mm}\\
$\Rightarrow$ {\bf \model:} $\quad\langle$ Event Level: \colorbox{green!60}{Moderate} $\rangle\quad\langle$ Effect Level: \colorbox{red!60}{Moderate} $\rangle\quad\langle$ Requirement Level: \colorbox{green!60}{Well-described}\\
$\langle$ {\bf Event:} Can you elaborate more on the Adderall? $\rangle$ ; $\langle$ {\bf Effect:} Can you elaborate more on how you feel after taking Adderall? $\rangle$ ; $\langle$ {\bf Req:} n/a $\rangle$ \\
\bottomrule
\end{tabular}}
\caption{An error analysis of \model. \textcolor{magenta}{{\em Event}}, \textcolor{blue}{{\em Effect}}, and \textcolor{black!30!cyan}{{\em Requirement}} spans are illustrated in magenta, blue, and cyan, respectively. [Best viewed in color]. More examples in Appendix~\ref{appx:additional analysis}.
}
\label{tab:analysis}
\end{table*}

\paragraph{Ablation Study.}
Our ablation study assesses the contribution of two major modules in \model: verifier (reward model) and taxonomy. We present experimental results in Table \ref{tab:ablation}. We observe that \model\ suffers a minute loss in performance when the verifier (Rew) is removed, with performance not getting impacted quantitatively. This raises questions on the impact of reward modeling in \model. However, qualitative analysis and human evaluation (Section \ref{sec:qualityError}) reveal substantial improvements in generation quality with reward modeling. On the other hand, when removing both verifier and taxonomy, we see a significant decline in the model's performance, showing the essence of taxonomy in the whole generative framework. For instance, R1 score declines from from $89.30$ to $68.20$, and BERTScore (F1) drops from $98.74$ to $95.48$. \mar{Surprisingly, Mistral$+$\tx{} yields performance comparable to \model. Later, on qualitative analysis (Appendix \ref{appx:mistralsamples}), we observed limitations in the generations of Mistral, especially pertaining to the context and LLM’s understanding of the relation between support attributes.}

In general, all language models in our study see a significant jump in performance with the addition of our modules (Table~\ref{tab:ablation}). For instance, if we consider BERTScore (F1), then we observe improvement of $+2.46$ points, $+3.00$ points, and $+2.55$ points on Mistral, Phi-3, and Llama-3, respectively, on the addition of both modules.

\paragraph{Qualitative and Error Analysis.}
\label{sec:qualityError}
To further assess the quality of generations, we perform a qualitative analysis of the model's generation. Also, we aim to justify the significance of \model's reward module, {\em verifier}, despite observing only a small improvement in the quantitative metrics. We present a sample instance from our framework in Table \ref{tab:analysis}. In this case, \model{} is able to identify that the {\em event} attribute is moderately present and can be further improved. On the other hand, the {\em effect} and {\em requirement} questions are direct and more aligned with the context. Evaluators attest that there exist cases, where \model\ performs better than the gold standard.

To further assess the credibility of the model's performance and metric selection, we show additional examples and analyze them in Table \ref{tab:without_tax_samples} (c.f. Section \ref{app:additionalAnalysis}; Appendix). Although BLEU and ROUGE may reward lexical similarity (e.g., from template structure), \model\ consistently improves semantic relevance by steering the model to engage with under-addressed or missing entity types. This supports our broader claim that the MH-Copilot pipeline, guided by taxonomy-informed prompts, enhances generation quality.

\paragraph{Human Evaluation.} Findings from Table \ref{tab:ablation} show that adding the verifier yields only a slight boost to the performance, questioning reward modeling’s impact. Although verifier  modestly improves generation quality, we conduct a human evaluation to more comprehensively analyze it. The aim is to assess the generic \model's generation as well as compares it with {\em non-verifier} variant. Our human evaluation involves scoring the \model's generation based on three domain-centric parameters and four linguistic parameters. Domain-centric evaluation metrics are defined as follows: {\bf (D1) contextual relevance},{\bf (D2) empathy}, {\bf (D3) value addition}. On the other hand, linguistic evaluation metrics are defined as follows: {\bf (L1) relevance}, {\bf  (L2) coherence}, {\bf (L3) fluency}, {\bf (L4) consistency}. Definitions of each parameter is in Appendix \ref{appx:defintions}. We curated a sample of $40$ random instances, and we asked evaluators to rate linguistic metrics (L-1,2,3,4) on a scale of $1$ (worst) to $5$ (best) and domain-centric metrics (D-1,2,3) on a scale of $1$ to $3$. Evaluators are from mixed professions, aged between 21-35, and have been previous users of mental health subreddit channels. We repeat the same process for {non-verifier} setup to add justification for the relevance of reward modeling from a qualitative aspect.

As shown in Table \ref{tab:humaneval}, the Domain section presents domain-centric results, whereas the Linguistic section presents linguistic results. Evidently, \model\ excels across all suites of domain-centric and linguistic metrics. \mar{On linguistic aspects, \model's generations are fluent ($4.02$) as well as coherent ($3.84$), as fluency and coherency receive maximum human feedback rating after adding the {\em verifier} module.} This further indicates that the {\em verifier} module outputs outperform the {\em non-verifier} setup, showing qualitative superiority. Additionally, annotators further state that the generations from \model, in some cases, surpass the gold standard.

\begin{table}[t]\centering
\scriptsize
\begin{tabular}{lccccccc}
\toprule
& \multicolumn{3}{c}{\bf Domain} & \multicolumn{4}{c}{\bf Linguistic}\\ \cmidrule(lr){2-4} \cmidrule(lr){5-8}
& \bf D1 & \bf D2 & \bf D3 & \bf L1 & \bf L2 & \bf L3 & \bf L4 \\  \midrule
\bf w/o Verifier & 3.27 & 1.82 & 2.19 & 3.46 & 3.70 & 3.82 & 3.80 \\ 
\bf w/ Verifier & \bf 3.43 & \bf 2.27 & \bf 3.31 & \bf 3.62 & \bf 3.84 & \bf 4.02 & \bf 3.89 \\ \bottomrule
\end{tabular}
\caption{Human evaluation on the responses generated from \model\  when compared to {\model} without verifier. We observe that the performance of \model\ across all metrics is better than without verifier module.}
\label{tab:humaneval}
\vspace{-3.5mm}
\end{table}

\section{Conclusion}
To address the persistent issue of low response rates on support seeker's posts in OMHCs, we proposed \model, a reinforcement-learning-based framework designed to assess and enhance the clarity of support-seeking posts. Our contributions include \data, a novel dataset comprising $4,760$ posts annotated with key support attributes event, effect, and requirement at both span and intensity levels. Additionally, we propose a dedicated taxonomy, \tx, for controlled prompting and generation. Next, we propose \model\ that employs \tx{} to generate contextually relevant and tailored questions, prompting users to provide more complete and actionable information to improve their posts inorder to elicit more responses from support-providers. Our extensive benchmarking against notable language models demonstrated consistent improvements across $11$ metrics. We concluded our work with an exhaustive analysis supplemented with human evaluation.  

\section{Ethical Considerations}
\mar{Our work acknowledges that support-seekers often face disparities in engagement, whereas those who articulate their issues clearly tend to receive responses, while others remain unnoticed. It is worth noting that our system is not intended to `diagnose' or `suggest' remedies for mental health issues. Its sole purpose is to help users articulate their thoughts more clearly, enabling a better peer support ecosystem. Our system is designed to be non-intrusive and supportive, which gradually encourages users in crisis to seek peer help.}

\section{Limitations and Future Scope} 
While \model\ demonstrates significant improvements in enhancing support-seeking posts within OMHCs, it has limitations, including reliance on the specific dataset and potential biases inherent in language models. Future work could explore expanding the dataset to include diverse online platforms and refining the model to better address nuances in user expression. Furthermore, \model\ is inspired by therapeutic principles and aligns with prior work on explainable and controllable generation in sensitive domains like mental health. The modularity also allows for extensibility to other domains. Additionally, integrating real-time feedback mechanisms and user-centered evaluations could further enhance the model’s applicability and effectiveness in diverse mental health contexts.

\section{Acknowledgement}
The authors acknowledge the support of the Infosys Foundation through CAI at IIIT-Delhi.

\bibliography{custom}

\newpage

\appendix
\noindent{\Large \textbf{Overview of Appendices}}
\begin{itemize}[leftmargin=*]
\item Appendix~\ref{appx:attributes}: Support Attributes.
\item Appendix~\ref{appx:additional rationale}: Additional Annotation Rationale.
\item Appendix~\ref{appx:annotator guidelines}: Annotator Guidelines.
\item Appendix~\ref{appx:subreddits}: Dataset Analysis.
\item Appendix~\ref{appx:prompts}: Prompts.
\item Appendix~\ref{appx:experimentDetails}: Training Details.
\item Appendix~\ref{appx:additional analysis}: Additional Examples.
\item Appendix~\ref{appx:mistralsamples}: Comparison of Mistral and \model.
\item Appendix~\ref{appx:defintions}: Human Evaluation Parameters.
\end{itemize}

\section{Support Attributes}\label{appx:attributes}

\paragraph{Examples.} Table \ref{tab:attrsamples} presents example text spans from Reddit posts illustrating each support attribute.


\begin{table*}[t]\centering
\centering
\scriptsize
\begin{tabular}{lp{53em}}
\toprule
\bf Attributes & \bf Definition \\
\midrule
\textcolor{magenta}{\bf Event} & Encapsulates the specific situation, activity, or event that is the focal point of the support seeker’s concern. The explicit detailing of such events provides a contextual background essential for overall background understanding, as suggested by \citet{sharma2020computational}. \textbf{Example:} {\em ``So I have been taking adderall 10mg for a little over a week and I’ve been testing it. Today I decided to take Adderall and coffee at the same time.''}\\
\textcolor{blue}{\bf Effect} & Targets the impact or consequences of the identified event from the support seeker. By elucidating the effect, the post conveys the emotional or practical repercussions of the situation, thereby inviting more targeted and empathetic responses. \newline  \textbf{Example:}  {\em ``I think this was a bad idea because I felt extremely jittery and anxious, but the worst part was I came down from them at the same time.''}\\

\textcolor{black!30!cyan}{\bf Requirement} & lays out the expectation (e.g., informational support, instrumental aid, etc.) of support seeker from peers. It is crucial in directing the nature of the assistance sought, and thereby guiding the potential support trajectory. This aligns with the insights laid out by \citet{sharma2022humanaicollaborationenablesempathic}, highlighting the importance of clearly articulated needs for effective support. \newline \textbf{Example:} {\em ``So if you guys got any tips that would be helpful, like if its the flu, or the Adderall, or caffeine, or anyone else has experienced this.''}\\ 
\bottomrule
\end{tabular}
\caption{Definition and examples of support attributes.}
\label{tab:attrsamples}
\end{table*}

\section{Additional Annotation Rationale} \label{appx:additional rationale}
In the annotation process, we are required to identify the precise support attributes that potentially appeal to support providers to respond. Considering our goal of proposing a solution that does not target just one metric and its improvement but a more holistic improvement in support-seeker experiences, we realize that generating high-quality suggestions for the users to improve their posts would require the rationales governing the annotator’s judgment for the presence or absence of the critical elements. 
Moreover, at times, support seekers touch upon the subject without providing adequate insights, prompting the support providers to respond in an abstract way. Therefore, we also mark the intensity/level of their presence in the post. Finally, to provide proactive support, we annotate reference questions that would encourage the user to write an appropriate post considering their current attribute-level response.

\section{Annotator Guidelines}\label{appx:annotator guidelines}
\mar{In this annotation task, identify and label spans within Reddit posts that express emotional distress, mental health symptoms, coping mechanisms, or recovery efforts. Focus on subjective, psychologically significant content such as feelings of sadness, anxiety, hopelessness, self-perception, mentions of therapy or medication, or descriptions of distressing experiences. For each span, assign an intensity score from 0 to 2. Avoid annotating spans that are purely narrative, background, or unrelated to mental health. If a span includes a long sentence expressing different attributes (event, effect or requirement), break the sentence using suitable punctuation and annotate them separately. Be consistent, base your judgment on the text itself, and flag any uncertain or potentially serious content for review if required. Include the title for annotation if the title includes a significant context.}

\section{Dataset Analysis}\label{appx:subreddits}

\begin{table*}[!h]
\centering
\scriptsize
\begin{tabular}{clp{30em}H} \toprule
\textbf{\#}     &  \textbf{Subreddit} & \textbf{Description} & \textbf{\# Posts} \\\cmidrule{1-4}
1 &r/Anxiety               & A subreddit for support and discussion around anxiety disorders.                 & 469 \\ 
2 & r/ptsd                 & A community for individuals dealing with post-traumatic stress disorder.         & 494 \\ 
3 & r/addiction            & Support and resources for individuals battling addiction.                        & 487 \\ 
4 &r/ADHD                  & A subreddit focused on attention-deficit/hyperactivity disorder discussions.     & 423 \\ 
5 & r/alcoholicsanonymous  & A forum for members of Alcoholics Anonymous and those interested.                & 498 \\ 
6 & r/Anger                & A community to discuss and manage anger issues.                                  & 464 \\ 
7 & r/BPD                  & A space for individuals with borderline personality disorder.                    & 519 \\ 
8 & r/depression           & Support and discussion for those dealing with depression.                        & 547 \\ 
9 & r/domesticviolence     & A community for survivors of domestic violence seeking support.                  & 425 \\ 
10 & r/getting\_over\_it   & A space for advice on overcoming challenges and struggles.                       & 476 \\ 
11 & r/mentalillness       & Discussions and support around various mental illnesses.                         & 484 \\ 
12 & r/OpiatesRecovery     & A subreddit for recovery from opiate addiction.                                  & 493 \\ 
13 & r/rapecounseling      & Support and counseling for survivors of sexual violence.                         & 481 \\ 
14 & r/sad                 & A community for sharing and coping with feelings of sadness.                     & 486 \\ 
15 & r/selfharm            & A supportive space for those dealing with self-harm.                             & 467 \\ 
16 & r/selfhelp            & Resources and discussions on self-improvement and help.                          & 419 \\ 
17 & r/socialanxiety       & A subreddit for individuals with social anxiety disorder.                        & 461   \\ 
\bottomrule
\end{tabular}
\caption{Description of the various subreddits used in the BeCOPE~\cite{criticalBehaviorAseem} and also the \data{} dataset.}
\label{tab:subreddits}
\end{table*}
Table \ref{tab:subreddits} presents a comprehensive list of the subreddits included in the \data{} dataset. These subreddits were carefully selected to provide a diverse and representative sample of online discourse across a wide range of topics and communities. The selection process aimed to capture variations in language use, sentiment, community norms, and topical focus, all of which are crucial for robust linguistic and social analysis.

 From the source BeCOPE dataset, we utilize interactive posts from $17$ subreddits, which constitutes a total of $4760$ posts. The final dataset, \data, contains post title, post body, annotated post body, event-intensity, effect-intensity, requirement-intensity, event-question, effect-question, and requirement-question. The average post length is $179.62$ words, and we observe a total of $2125$ posts in which the event is absent or moderately present, whereas 
$2781$ posts for effect and $2976$ posts for requirement, where we observe it to be absent or moderately present. Thus, a large percentage of the posts in the dataset either have a complete absence or only a moderate presence of the attributes. This indicates that the majority of support seekers are unable to properly express their problems. Our annotators' marked rationale's average event-span-length is $65.70$ words, $26.48$ words for effect-span-length, and $19.31$ words for requirement-span-length. This marks that the support seekers have mainly written about the triggering incidents but have provided very little information about the supporting effects of that incident and what they want as help from the community. The post-body length is observed to follow an increasing trend as the intensity levels increase for each cue.

\section{Prompts}\label{appx:prompts}
Table \ref{tab:prompts} describes the user and system prompts used for the Language Models for generating the guiding questions. The system prompt briefly describes the role of a Online Mental Health Platforms, support providers and support seekers. \\
The user prompt consists of the reddit post with event, effect and requirement spans as well as their respective intensity levels.

\begin{table*}[h]
\centering
\resizebox{\textwidth}{!}{%
\begin{tabular}{p{58em}} \toprule
{\centering \textbf{System Prompt}} \\\cmidrule{1-1}
{
A support seeker on a peer-to-peer (P2P) Online Mental Health Platform (OMHP) is an individual who utilizes digital services to seek assistance/help for managing and improving their mental health, typically through interactions with peer groups or self-help resources. \newline

The parameters are defined as follows: \newline
\textbf{Event:} This parameter encapsulates the specific situation, activity, or event that is the focal point of the support seeker’s concern. The explicit detailing of such events provides a contextual background essential for empathetic understanding. \newline
\textbf{Effect:} This aspect targets the impact or consequences of the identified event on the support seeker. By elucidating the effect, the post conveys the emotional or practical repercussions of the event, thereby inviting more targeted and empathetic responses. \newline
\textbf{Requirement:} This parameter is critical in directing the nature of the assistance sought. It ranges from emotional and informational support to instrumental aid, thereby guiding the potential response trajectory. \newline

In the posts on OMHP, these parameters can have intensity ranging from 0 to 2, where 0 means absent, 1 means present but needs clarification, and 2 being well-described based on the presence of these parameters in the post.

Consider the following post by a support seeker on a OMHP, in which the spans of text representing \textit{event}, \textit{effect}, and \textit{requirement} have been marked. Also, the intensity levels for each of the parameters in the post have been provided along with the post. The post is context of the victim.
The <es> and <ee> tags encapsulate the spans for the \textit{event} parameter, <efs> and <efe> tags encapsulate the spans for the \textit{effect} parameter, and <rs> and <re> tags encapsulate the spans for the \textit{requirement} parameter.
}
\\
\midrule
{\centering \textbf{User Prompt}} \\\cmidrule{1-1}
{\textbf{Post:} <Post Body> 

\textbf{Event scale:} \qquad \textbf{Effect scale:} \qquad \textbf{Requirement scale:}

\textbf{Schema:} \{event\_question: , effect\_question: , requirement\_question: \}

Generate 3 questions following the schema provided below the post, for helping the support giver to understand more about the victim. Strictly follow the question format of schema. Give only the json output as specified in the schema and no explanation needed.
}\\

\bottomrule
\end{tabular}
}

\caption{The prompts used by language models to generate the guiding questions. The system prompt is a description of the task and description of all the attributes. The user prompt contains the Reddit post, intensity scale and schema for the generation format.}

\label{tab:prompts}
\end{table*}

\section{Training Details} \label{appx:experimentDetails}
We fine-tune four LMs: LLaMA 3-instruct (8B), Phi-3-mini-4k-instruct (3.8B), Mistral-instruct-v0.1 (7B), and Gemma-2-instruct (2B) using the QLoRA \cite{qlora} method (quantization=8 bit, r=16, alpha=32, dropout=0.05). The context length was set to 1024 tokens, optimized for a balance between the model's limit and computational efficiency. Training is carried out on 1x NVIDIA A100 GPUs (80GB) and 1x RTX A6000 GPUs (50GB), ensuring sufficient capacity for gradient accumulation and model checkpoints. We use a learning rate of 2e-5 for SFT to achieve stable convergence with AdamW optimizer. The training was carried out for 2 epochs after which the models started overfitting.
We employed the BLEU, ROUGE and METEOR metrics with default parameters from the Hugging Face\footnote{https://huggingface.co/} Evaluate library\cite{evaluate} for our evaluations. All experiments were run under Python 3.8.19, with PyTorch 2.4.1 and Transformers 4.45.0\footnote{https://huggingface.co/docs/transformers}.

\section{Additional Examples}\label{appx:additional analysis}
Table \ref{tab:additional analysis} presents representative outputs from the full \model{} pipeline. For each support attribute, we display the extracted text spans, their predicted intensity scores, and an associated guiding question designed to probe that attribute.

\begin{table*}[t]
\centering
\scriptsize
\begin{tabular}{c p{58em}} \toprule
\textbf{\#} & {\centering \textbf{Summaries}} \\\cmidrule{1-2}

1. & {\bf OP:} {\em \textcolor{magenta}{So I have been taking adderall 10mg for a little over a week and I've been testing it. Today i decided to take adderall and coffee at the same time.} \textcolor{blue}{I think this was a bad idea because i felt extremely jittery and anxious, but the worst part was i came down from them at the same time. I am not fully sure of my personal side effects of adderall yet but coming down at the same time from both I had this weird feeling of nausea, weak, bad headache, tired, extremely zoned out, and almost like a not knowing what's going on kinda feeling, also i'm just getting over the flu. Not sure why this happened but it sucked.} \textcolor{black!30!cyan}{Should I just avoid caffeine for now on, I can do that cause the adderall is doing way more than the caffeine was.} \textcolor{magenta}{The reason i'm asking is cause I couldn't find anyone having this problem.} \textcolor{black!30!cyan}{
So if you guys got any tips that would be helpful, like if its the flu, or the adderall, or the caffeine, or if anyone else has experienced this.}} \vspace{2mm}\\
& $\Rightarrow$ {\bf Annotated:} $\quad \langle$ Event Level: Moderate $\rangle \quad \langle$ Effect Level: Well-described $\rangle \quad \langle$ Requirement Level: Well-described $\rangle$\\
& $\langle$ {\bf Event:} Can you elaborate more on why you are taking adderall? $\rangle$ ; $\langle$ {\bf Effect:} n/a $\rangle$ ; $\langle$ {\bf Requirement:} n/a $\rangle$ \vspace{2mm}\\
& $\Rightarrow$ {\bf \model:} $\quad$ Event Level: \colorbox{green!60}{Moderate} $\quad$ Effect Level: \colorbox{red!60}{Moderate} $\quad$ Requirement Level: \colorbox{green!60}{Well-described} \\ 
& $\langle$ {\bf Event:} Can you elaborate more on the Adderall? $\rangle$ ; $\langle$ {\bf Effect:} Can you elaborate more on how you feel after taking Adderall? $\rangle$ ; $\langle$ {\bf Req:} n/a $\rangle$ \\ \midrule

2. & {\bf OP:} {\em Long story short - 

\textcolor{black!30!cyan}{Am I completely bonkers here?} \textcolor{magenta}{I get that high energy exercise is good for improving dopamine and norepinephrine  recommended, which is for several reasons but one big one is I get winded climbing short staircases. I'm 250lbs and haven't even walked for exercise in 6 months so I feel it's a terrible idea to jump directly into HIIT. }

\textcolor{black!30!cyan}{Am I completely bonkers here?} \textcolor{magenta}{I get that high energy exercise is good for improving dopamine and norepinephrine levels and have had good results from interval training in the past, but I also don't want to injure myself or trigger a heart attack. I have promised to do Yoga for a while, starting slow, and then looking into HIIT later but he called me lazy and said I'm hiding behind my diagnosis to avoid changing bad habits. :(  }

\textcolor{magenta}{He's not usually so mean, but he's been very pushy about this. I don't know what to do.}} \vspace{2mm}\\
& $\Rightarrow$ {\bf Annotated:} $\quad \langle$ Event Level: Well-described $\rangle \quad \langle$ Effect Level: Absent $\rangle \quad \langle$ Requirement Level: Moderate $\rangle$\\
& $\langle$ {\bf Event:} n/a $\rangle$ ; $\langle$ {\bf Effect:} How did doing yoga make you feel? $\rangle$ ; $\langle$ {\bf Req:} Can you elaborate more on what can help you change your bad habits? $\rangle$ \vspace{2mm}\\
& $\Rightarrow$ {\bf \model:} \quad Event Level: \colorbox{red!60}{Moderate} \quad Effect Level: \colorbox{red!60}{Moderate} \quad Requirement Level: \colorbox{red!60}{Absent} \\
& $\langle$ {\bf Event:} Can you elaborate more on your diagnosis? $\rangle$ ; $\langle$ {\bf Event:} Can you elaborate more on your diagnosis ? $\rangle$ ; $\langle$ {\bf Event:} Can you elaborate more on your diagnosis ? $\rangle$\\
\midrule

3. & 
\textbf{OP}: Hey everyone, I'd really appreciate some assistance with this one so please stick with me.

A bit of back ground;
\textcolor{magenta}{I'm a 25 year old female, I've been diagnosed with BPD before someone else diagnosed me with Bipolar type 2. I've always have massive issues with anxiety and recently OCD, i also had some self harming tendencies.}

But one thing really bugs me. \textcolor{magenta}{Since I was a young teen I've had these extremely in depth fantasys about being someone else, not that odd I know, but it's to a crazy extent.} 
\textcolor{blue}{I've never brought it up because honestly  I'm embarrassed}.\textcolor{magenta}{ Always about being this one person I've created in my head, sometimes I fantasize about being them at differnt stages of their life.<es> But always the same person.<es> I have created all these imaginary friends and relationship that this persons encounters. My brain acts out full conversations and everything....}

\textcolor{magenta}{I thought maybe this had begun as a coping mechanism after being so alone for so long when I was younger.(Btw were talking about living in these fantasies for multiple hours a day, while I work, while I drive, sometimes I can't even concentrate on a movie because of them) it's all contained in my head and no one knows it's happening.}

\textcolor{magenta}{But sometimes I'll catch myself making a facial expression/ sound or movement that's related to whats happening in my head not real life...}

\textcolor{blue}{I have friends now and life's okay, so I don't know why I still do this, I'm frightened there may be something wrong with my brain that's laying dormant... }

\textcolor{cyan}{So does anyone have a name for what this may be?}

If you read all of this, thank you so much, really. \\
& $\Rightarrow$ {\bf Annotated:} $\quad \langle$ Event Level: Well-described $\rangle \quad \langle$ Effect Level: Moderate $\rangle \quad \langle$ Requirement Level: Moderate $\rangle$\\
& $\langle$ {\bf Event:} n/a $\rangle$ ; $\langle$ {\bf Effect:} Can you elaborate more on how the fantasies make you feel? $\rangle$ ; $\langle$ {\bf Requirement:} Can you elaborate more on what would help you overcome these delusions ? $\rangle$ \vspace{2mm}\\
& $\Rightarrow$ {\bf \model:} $\quad$ Event Level: \colorbox{red!60}{Moderate} $\quad$ Effect Level: \colorbox{green!60}{Moderate} $\quad$ Requirement Level: \colorbox{red!60}{Absent} \\ 
& $\langle$ {\bf Event:} Can you elaborate more on your fantasies? $\rangle$ ; $\langle$ {\bf Effect:} Can you elaborate more on how you feel about your fantasies ?$\rangle$ ; $\langle$ {\bf Req:} What do you need help with now that you are having fantasies about being someone  $\rangle$ \\

\bottomrule
\end{tabular}
\caption{[Best viewed in color] A comparative analysis of positive and error cases of \model. Case \#1 and Case \#3 demonstrate the ability to capture insightful nuances, whereas Case \#2 demonstrates erroneous generation escalated from misclassification. \textcolor{magenta}{{\em Event} is illustrated in magenta}, \textcolor{blue}{{\em effect} spans in blue}, and \textcolor{black!30!cyan}{{\em requirement} spans in cyan.}
Additionally \textcolor{green!60}{Correct attribute intensity are shown in green} and \textcolor{red!60}{incorrect attribute intensity in red.}}

\label{tab:additional analysis}
\end{table*}

\section{Comparison of Mistral and \model}\label{appx:mistralsamples}
Table \ref{tab:mistralsamples} shows a comparative study between the generations of Mistral and \model. It clearly shows that RLHF improved the quality of the generated guiding questions.
In case \#1, \model{} is able to identify that the event attribute is moderately present and can be further improved. On the other hand, the effect and requirement questions are direct and 
more aligned with the context.\\
Similarly, case \#2 shows that the support seeker had already described much about the effect attribute on the topic related to `adderall' but \model\ has incorrectly classified its intensity and generated a question under that category. In this case, the fluency of the event question has also suffered. This proves that the model's performance remains intact till the pipeline's other modules remain intact, as it's evident that errors at the early phase get escalated to subsequent steps.

\begin{table*}[!t]
\centering
\scriptsize
\begin{tabular}{cp{62em}} \toprule
\textbf{\#} & {\centering \textbf{Reddit Posts}} \\ \midrule
1. & {I keep having positive dreams about the person that raped me over a year ago. he has verbally abused me in public for the past year and made my life a misery. I am terrified of him. why do i keep having dreams about him where he’s being nice to me and apologising for what he did, in my dreams he is a nice person but in real life he’s not. please help me understand why this is happening. \hfill [\textcolor{magenta}{\bf Event:} 2; \textcolor{blue}{\bf Effect:} 1; \textcolor{black!30!cyan}{\bf Requirement:} 2]} \\ \cmidrule{2-2}
& \textbf{Mistral+\tx{}:} \\
& \hspace{5mm} \textit{Effect question}: Can you elaborate more on how you feel about the dreams? \\ \cmidrule{2-2}
& \textbf{\model{}:} \\ 
& \hspace{5mm} \textit{Effect question}: Can you elaborate more on how the dreams make you feel ? \\ \midrule

2. & {I loved the Psych Ward I felt so safe and surrounded by nice non-judgemental people but of course my anxiety ruined it and it just became torture I felt on the verge of throwing up 24/7 and I barely left my room and couldn’t eat and every time I went for meds I shook like I had Parkinsons because my dumb brain can’t handle people focusing on me They were so nice man they even covered the mirrors for my BDD I wanna go back so bad but I’ll feel bad there because I feel bad everywhere and I just wanna die so bad. It’s a loss loss I can’t win this battle \hfill [\textcolor{magenta}{\bf Event:} 1; \textcolor{blue}{\bf Effect:} 2; \textcolor{black!30!cyan}{\bf Requirement:} 0]} \\ \cmidrule{2-2}
& \textbf{Mistral+\tx{}:} \\
& \hspace{5mm} \textit{Event question}: Can you elaborate more on what caused your anxiety? \\ 
& \hspace{5mm} \textit{Requirement question}: What do you need help with now that you feel bad everywhere? \\ \cmidrule{2-2}
& \textbf{\model{}:} \\ 
& \hspace{5mm} \textit{Event question}: Can you elaborate more on why you were in the psych ward? \\ 
& \hspace{5mm} \textit{Requirement question}: What do you need help with now that you feel you were tortured in the psych ward? \\ \midrule

3. & {So I have been taking adderall 10mg for a little over a week and I\'ve been testing it.Today i decided to take adderall and coffee at the same time.I think this was a bad idea because i felt extremely jittery and anxious, but the worst part was i came down from them at the same time.I am not fully sure of my personal side effects of adderall yet but coming down at the same time from both I had this weird feeling of nausea, weak, bad headache, tired, extremely zoned out, and almost like a not knowing what\'s going on kinda feeling, also i\'m just getting over the flu.Not sure why this happened but it sucked.Should I just avoid caffeine for now on, I can do that cause the adderall is doing way more than the caffeine was. The reason i\'m asking is cause I couldn\'t find anyone having this problem. So if you guys got any tips that would be helpful, like if its the flu, or the adderall, or the caffeine, or if anyone else has experienced this.  \hfill [\textcolor{magenta}{\bf Event:} 1; \textcolor{blue}{\bf Effect:} 2; \textcolor{black!30!cyan}{\bf Requirement:} 2]} \\ \cmidrule{2-2}
& \textbf{Mistral+\tx{}:} \\
& \hspace{5mm} \textit{Event question}: Can you elaborate more on why you took adderall and coffee at the same time ? \\ \cmidrule{2-2}
& \textbf{\model{}:} \\ 
& \hspace{5mm} \textit{Event question}: Can you elaborate more on why you are taking adderall? \\
\bottomrule
\end{tabular}
\caption{A comparative analysis of guiding questions generated by Mistral + \tx{} and \model{}.}
\label{tab:mistralsamples}
\end{table*}

\section{Human Evaluation Parameters} \label{appx:defintions}

Domain-centric evaluation metrics are defined as follows:
\begin{itemize}
    \item[\bf D1:] \textbf{Contextual relevance} of the generated questions with regard to the given input post; 
    \item[\bf D2:] \textbf{Empathy} shows how well the generated questions avoid triggering the support seeker;
    \item[\bf D3:] \textbf{Value addition} justifies how much is the generated question's response likely to add more value to the original post.
\end{itemize}
 On the other hand, linguistic evaluation metrics are defined as follows: 
 \begin{itemize}
     \item[\bf L1:] \textbf{Relevance} measures the selection of relevant content considering the reference utterance.     \item[\bf L2:] \textbf{Coherence} examines the structure and organization of the generated responses;
\item[\bf L3:] \textbf{Fluency} assesses the linguistic quality of the generated responses;
     \item[\bf L4:] \textbf{Consistency} evaluates factual alignment with the source utterance.
 \end{itemize}

\section{Additional Analysis} 
\label{app:additionalAnalysis}
To further assess the credibility of the selection of metrics and the model's performance, we perform additional qualitative analysis. Table \ref{tab:without_tax_samples} shows supporting examples for this analysis. 

\begin{itemize}[leftmargin=*]
    \item In the first example from our dataset, the Effect and Requirement entities are already well described (Level 2), whereas the Event entity is only minimally present (Level 1). Without CueTaxo, the model generates a follow-up targeting the Effect entity, already well-formed, thus missing an opportunity to probe the underdeveloped Event. With CueTaxo, the follow-up instead targets the Event, resulting in a more contextually useful and balanced question. This shift corresponds to an increase in both BLEU and ROUGE scores, aligning with improved semantic targeting.

    \item In the second example, the Requirement entity is completely missing from the post. Without CueTaxo, the model attempts to generate a follow-up targeting only this missing entity, while ignoring the Event (present at Level 1). CueTaxo instead guides the model to generate follow-ups for both missing and underdeveloped entities, better capturing the post's intent and producing a more complete response, again reflected in improved scores.

    \item In the third example, the question generated without CueTaxo is appropriate but fails to include any question related to the Effect entity. The CueTaxo-guided output includes this additional, contextually relevant query. Even though both versions may receive high BLEU/ROUGE due to templated phrasing, the CueTaxo version provides greater utility and completeness, contributing to a genuine gain in evaluation metrics.
\end{itemize}

\begin{table*}[!t]
\centering
\scriptsize
\begin{tabular}{cp{62em}} \toprule
\textbf{\#} & {\centering \textbf{Reddit Posts}} \\ \midrule
1. & {My partner can get very angry at times and it makes me feel uncomfortable. We now have a baby and I do not want him to be that way around her but I don’t know the best way to talk to him about this. What should I do if he ever is angry around her? I want him to know that it is absolutely unacceptable if he ever is like that. Any help would be really appreciated.
\hfill} \\ \cmidrule{2-2}
& \textbf{Without \tx{}:} \\
& \hspace{5mm} \textit{Event Question}: \\
& \hspace{5mm} 
\textit{Effect Question}: How did your partner's anger make you feel? \\ 
& \hspace{5mm} 
\textit{Requirement Question}:\\ \cmidrule{2-2}
& \textbf{With \tx{}} \\ 
& \hspace{5mm} 
\textit{Event Question}: Can you elaborate more on your partner's anger? \\ 
& \hspace{5mm}          \textit{Effect Question}:  \\ 
& \hspace{5mm}      \textit{Requirement Question}: \\ \midrule

2. & {I loved the Psych Ward. I felt so safe and surrounded by nice non-judgemental people, but of course, my anxiety ruined it and it just became torture. I felt on the verge of throwing up 24/7 and I barely left my room and couldn’t eat. Every time I went for meds I shook like I had Parkinson’s because my brain can’t handle people focusing on me. They were so nice—they even covered the mirrors for my BDD. I wanna go back so bad but I’ll feel bad there, because I feel bad everywhere. I just wanna die so bad. It’s a loss-loss. I can’t win this battle. \hfill } \\ \cmidrule{2-2}
& \textbf{Without \tx{}:} \\
& \hspace{5mm} 
\textit{Event Question}: \\ 
& \hspace{5mm}    
\textit{Effect Question}: \\ 
& \hspace{5mm}   \textit{Requirement Question}: What do you need help with now that you feel like you are losing control of your mental health? \\ 
\cmidrule{2-2}
& \textbf{With \tx{}:} \\ 
& \hspace{5mm} \textit{Event Question}: Can you elaborate more on why you were in the psych ward? \\ 
& \hspace{5mm}            \textit{Effect Question}: \\ 
& \hspace{5mm}    \textit{Requirement Question}: What do you need help with now that you feel you were tortured in the psych ward? \\ \midrule

3. & {I'll start off by saying some days I feel somewhat normal. But mainly I isolate myself from people and find other people annoying and sensitive... I also tend to (and like to) take things quite literally, perhaps due to the structure of programming. My apologies if this is the wrong sub for this, and my behavior described is actually quite common.  \hfill } \\ \cmidrule{2-2}
& \textbf{Without \tx{}:} \\
& \hspace{5mm} \textit{Event Question}:    \\
& \hspace{5mm}      \textit{Effect Question}:    \\
& \hspace{5mm}
\textit{Requirement Question}: What do you need help with now that you feel isolated from people? \\ \cmidrule{2-2}
& \textbf{With \tx{}:} \\ 
& \hspace{5mm} \textit{Event Question}:    \\
& \hspace{5mm}  
\textit{Effect Question}: Can you elaborate more on how you feel about socializing?     \\
& \hspace{5mm}  
\textit{Requirement Question}: What do you need help with now that you feel isolated from people? \\
\bottomrule
\end{tabular}
\caption{A comparative analysis of guiding questions generated without \tx\ and with \tx\ with Gemma.}
\label{tab:without_tax_samples}
\end{table*}
\end{document}